\documentclass[conference]{IEEEtran}
\IEEEoverridecommandlockouts
\usepackage{cite}
\usepackage{amsmath,amssymb,amsfonts}
\usepackage{algorithmic}
\usepackage{graphicx}
\usepackage{textcomp}
\usepackage{xcolor}
\usepackage{url}
\usepackage[bookmarks=false]{hyperref}

\usepackage{tikz}
\usepackage{textcomp}
\usepackage{hyperref}
\usepackage{lipsum}

\newcommand\copyrighttext{%
  \footnotesize \textcopyright 2019 IEEE. Personal use of this material is permitted. Permission from IEEE must be obtained for all other uses, in any current or future media, including reprinting/republishing this material for advertising or promotional purposes, creating new collective works, for resale or redistribution to servers or lists, or reuse of any copyrighted component of this work in other works.
  DOI: \href{https://doi.org/10.1109/AIKE.2019.00044}{10.1109/AIKE.2019.00044}}
\newcommand\copyrightnotice{%
\begin{tikzpicture}[remember picture,overlay]
\node[anchor=south,yshift=10pt] at (current page.south) {\fbox{\parbox{\dimexpr\textwidth-\fboxsep-\fboxrule\relax}{\copyrighttext}}};
\end{tikzpicture}%
}
    
\begin{document}

\title{Empirical comparison between autoencoders and traditional dimensionality reduction methods\\
}

\author{\IEEEauthorblockN{Quentin Fournier}
\IEEEauthorblockA{
Ecole Polytechnique Montreal\\
Montreal, Quebec H3T 1J4 \\
quentin.fournier@polymtl.ca}
\and
\IEEEauthorblockN{Daniel Aloise}
\IEEEauthorblockA{Ecole Polytechnique Montreal\\
Montreal, Quebec H3T 1J4 \\
daniel.aloise@polymtl.ca}
}

\maketitle

%
\copyrightnotice

\begin{abstract}
In order to process efficiently ever-higher dimensional data such as images, sentences, or audio recordings, one needs to find a proper way to reduce the dimensionality of such data. In this regard, SVD-based methods including PCA and Isomap have been extensively used. Recently, a neural network alternative called autoencoder has been proposed and is often preferred for its higher flexibility. This work aims to show that PCA is still a relevant technique for dimensionality reduction in the context of classification. To this purpose, we evaluated the performance of PCA compared to Isomap, a deep autoencoder, and a variational autoencoder. Experiments were conducted on three commonly used image datasets: MNIST, Fashion-MNIST, and CIFAR-10. The four different dimensionality reduction techniques were separately employed on each dataset to project data into a low-dimensional space. Then a $k$-NN classifier was trained on each projection with a cross-validated random search over the number of neighbours. Interestingly, our experiments revealed that $k$-NN achieved comparable accuracy on PCA and both autoencoders’ projections provided a big enough dimension. However, PCA computation time was two orders of magnitude faster than its neural network counterparts.
\end{abstract}


\section{Introduction}

In recent years, large sets of ever-higher dimensional data such as images~\cite{Krizhevsky09learningmultiple, lecun-mnisthandwrittendigit-2010, xiao2017/online}, sentences, or audio recordings have become the norm. One of the main problems that arise with such data is called the {\it curse of dimensionality}. Since the volume of the data space increase exponentially fast with the dimension, the data becomes sparse. Another problem is that distances computation time generally grows linearly with the dimension.

In order to process such high-dimensional data, one can learn a projection into a lower-dimensional space. Singular value decomposition (SVD) based methods such as principal component analysis (PCA) have been extensively used in that regard.

In 2006, Hinton and Salakhutdinov~\cite{HintonSalakhutdinov2006b} noted that a deep learning approach could be applied to dimensionality reduction using neural networks that learn to predict their input. Such networks are called \textit{autoencoders} and, once trained, yield a non-linear dimensionality reduction that outperforms SVD-based methods.

However, neural networks require more computation time or resources to be trained than most SVD-based methods. Therefore, neural networks may not be suitable for applications whose resources are limited such as prototyping. 

In this paper, we provide an analysis on how SVD-based methods compare with neural networks in image classification tasks. Three reference image datasets were used, namely MNIST, Fashion-MNIST, and CIFAR-10. Two SVD-based methods and two autoencoders were applied to reduce the dimension of each dataset and a $k$-NN classifier was trained on each projection. This work brings an insight on how much autoencoders trade computation time off for projection's quality as evaluated by the accuracy of a $k$-NN classifier. The same analysis as evaluated by the logistic regression and the quadratic discriminant analysis can be found in the supplementary material\footnote{\url{https://github.com/qfournier/dimensionality_reduction/blob/master/supplementary_material.pdf}}. To the best of our knowledge, this is the first work which addresses the relevancy of PCA compared to autoencoders in the context of image classification.

The rest of this article is organized as follows: Section \ref{projection_methods} describes our proposed approach in more detail. Section \ref{experimentation} presents the experimental results. Finally, Section \ref{conclusion} summarizes this work.

\section{Projection Methods}
\label{projection_methods}



SVD-based methods and neural networks are two fundamentally different approaches to dimensionality reduction. The former are exact and deterministic, whereas the latter settle for a small value of an objective function and are non-deterministic between training runs.

\subsection{SVD-based}

In order to have a fair comparison, a simple linear method called PCA and a more complex non-linear one called Isomap were used.

PCA finds the linear projection that best preserves the variance measured in the input space~\cite{DBLP:journals/corr/Shlens14}. This is done by constructing the set of data's eigenvectors and sorting them by their eigenvalues. Dimensionality reduction is done by projecting the data along the first $k$ eigenvectors, where $k$ is the dimension of the reduced space. Although PCA is both easy to use and very efficient, its effectiveness is limited when data is not linearly correlated.

The isometric feature mapping (Isomap)~\cite{tenenbaum_global_2000} is based on classical multidimensional scaling (MDS), which finds a non-linear embedding that best preserve inter-point distances. However, instead of computing the matrix of inter-point distances in one step, Isomap first compute the distance between each point and its k-nearest neighbours, and constructs a weighted graph $\mathcal{G}$. Inter-point distances are then evaluated by finding the shortest path in $\mathcal{G}$. Finally, classical MDS is applied to the distance matrix. By computing interpoint distances iteratively, Isomap is able to learn a non-linear embedding that preserve the intrinsic geometry of the data.

\subsection{Neural Networks}

Let us first introduce the framework of an autoencoder. Given an input $x$, its projection $z$, and its reconstruction $x'$, an autoencoder is composed of two networks:

\begin{itemize}
   \item[--] An encoder defined by the function $f(x)=z$ such that $x$ is the input and $z$ is the output of the network.
   \item[--] A decoder defined by the function $g(z)=x'$ such that $z$ is the input and $x'$ is the output of the network.
\end{itemize}

The training objective is to minimize the distance between the input $x$ and its reconstruction $g(f(x))=x'$. It is necessary to limit the capacity of the model to copy its input on its output in order to force the autoencoder to extract useful properties. One can impose a regularization term or limit the dimension of the projection $z$.

\begin{figure}[!t]
\centering
\includegraphics[height=95pt]{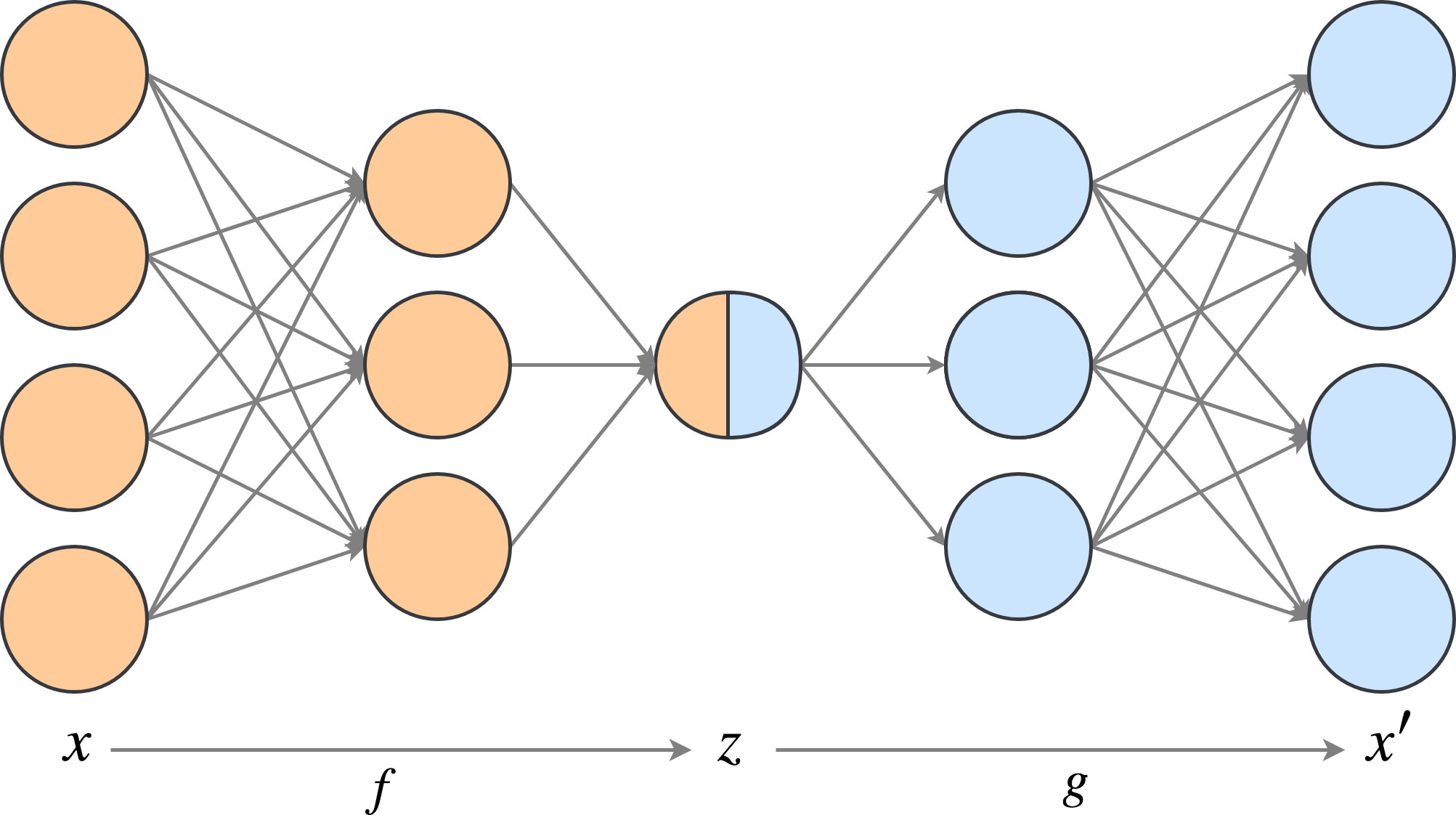}
\caption{An example of a deep autoencoder. The encoder is in orange and the decoder in blue. The 1-dimension hidden representation is the bicoloured node.}
\label{ae}
\end{figure}

\begin{figure}[!t]
\centering
\includegraphics[height=95pt]{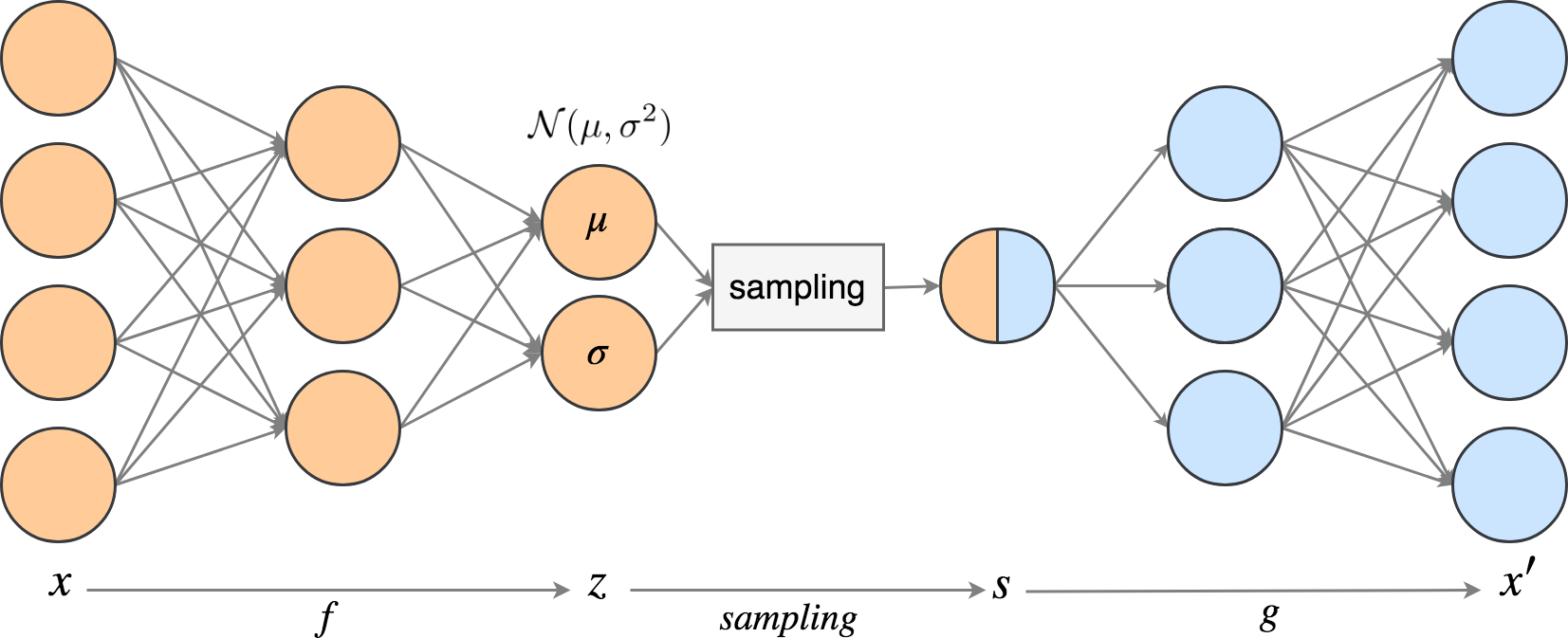}
\caption{An example of a variational autoencoder. The encoder is in orange and the decoder in blue. Note that the hidden representation $z$ is a density from which a value $s$ is sampled.}
\label{vae}
\end{figure}

Hereafter, we will use two autoencoders: a deep autoencoder (DAE, \figurename \ref{ae}) and a variational autoencoder (VAE, \figurename \ref{vae}). The former is a standard network whose encoder and decoder are multilayer perceptrons. The latter is a generative model first introduced in 2013 by Kingma et al.~\cite{journals/corr/KingmaW13} and differs by its latent space $z$ being non-deterministic. Hence, VAEs can both generate new samples and provide a probabilistic low-dimensional projection. Given the model's parameters $\theta$, the projection of $x$ into $z$ is a Gaussian probability density noted $q_{\theta}(z|x)$. This prior knowledge is embedded into the objective function through the following regularization term $KL(q_{\theta}(z|x)||p(z))$ which is Kullback-Leibler divergence between the encoder distribution and the expected distribution $p(z)\sim \mathcal{N}(0,1)$.

Although autoencoders can be specialized to some type of input data (\textit{e.g.}, by adding convolutions for the processing of images), we decided against to keep the comparison as fair as possible.

When applied to dimensionality reduction, the autoencoder is trained with both the encoder and decoder. Then the decoder is discarded and the output of the encoder is treated as the data's projection. This approach yields a non-linear generalization of PCA provided that both the encoder and decoder have at least one hidden layer~\cite{HintonSalakhutdinov2006b}. Note that for the VAE, the sampled vector $s$ is used.


\section{Experiments}
\label{experimentation}



Let us describe the experimental framework: each dataset has already been fairly split into a training set and a test set. In the case of autoencoders, a tenth of the training set is used as a validation set to tune the hyperparameters and to evaluate the stopping criterion during training. A $k$-nearest neighbours ($k$-NN) classifier is trained on each learned projection of the training set, with a cross-validated random search over $k$. Finally, the test set is projected by each method and classified by the associated $k$-NN. We investigate the accuracy of the $k$-NN algorithm for different projection dimensions.

\subsection{Datasets}

Methods described in section \ref{projection_methods} can be applied to any vector-represented data. In this study, we focus on small images. We used the well-known database of handwritten digits MNIST~\cite{lecun-mnisthandwrittendigit-2010} and its modern fashion analogue called Fashion-MNIST~\cite{xiao2017/online}. Both datasets consist of $(28\times28)$ grayscale images divided as 60,000 training and 10,000 testing examples. In addition, we used CIFAR-10~\cite{Krizhevsky09learningmultiple} which is composed of $(32\times32)$ coloured images divided as 50,000 training examples and 10,000 test examples. All three datasets contain ten equally distributed classes. Images were normalized to speed up the training and were treated as 1-dimensional vectors of sizes 784 and 3072, respectively.



\subsection{Detailed Architectures and Parameter Settings}

Experiments were implemented in Python using \textit{Keras} and \textit{scikit-learn} libraries. Training times were computed on a remote server equipped with two 16-core processors and two Titan Xp. Please note that neural networks training times depend highly on the graphics card used. The code is available on Github\footnote{\url{https://github.com/qfournier/dimensionality_reduction}}.

Encoders were five layers deep with each layer half the size of the previous one. The projection's size is a parameter that ranges from 1 to 99. Batch normalization was applied before each activation function. The non-linearity of choice was the Rectified Linear Unit (ReLU) except for the last layer which used a linear activation function. Decoders were symmetric with the exception that the output layer's activation function was the sigmoid, providing better results due to the input being normalized. Adam~\cite{DBLP:journals/corr/KingmaB14} optimizer was used as it offers both fast training and good generalization performance. Finally, early stopping was applied after a patience of 10 iterations. Weights were initialized based on the uniform distribution suggested by Glorot and Bengio~\cite{pmlr-v9-glorot10a}.

The number of neighbours $k$ was subject to a random search over 60 different $k \in [1, \sqrt{n}]$ with $n$ the size of the training set. The test accuracy was estimated with a 5-fold cross-validation.

Isomap complexity is $O(n^3)$~\cite{DBLP:journals/iacr/OuSWZC17} with $n$ is the number of examples. In order to make the experiments feasible, this method was trained on a sample of size 10,000\footnote{Note that it would take around $6^3=216$ and $5^3=125$ longer to train Isomap with MNIST and CIFAR full training set, respectively.}. Empirically, we found that more training examples didn't offer any substantial gains compared to the computational cost.

\subsection{Results}

Neural networks are non-deterministic as they are trained with random batches of examples and their weights are randomly initialized. In addition, all projection methods are evaluated with a $k$-NN classifier whose hyperparameter is selected through a random search. Finally, computation times depend to some degree on various factors out of our control. In order to take these random factors into account, experiments were repeated 5 times and the mean with the standard deviation is reported.

\subsubsection{Accuracy}

\begin{table}[]
\caption{Highest $k$-NN accuracy $\pm$ standard deviation (in \%).}
\label{acc_table}
\begin{tabular}{l|ccc}
       & MNIST                  & Fashion-MNIST          & CIFAR-10               \\ \hline
PCA    & $97.48 \pm 0.08$        & $85.56 \pm 0.05$          & $42.35 \pm 0.23$          \\
ISOMAP & $94.87 \pm 0.16$        & $81.17 \pm 0.22$          & $33.61 \pm 0.52$          \\
DAE    & \textbf{$97.82 \pm 0.08$} & \textbf{$87.62 \pm 0.11$} & \textbf{$45.77 \pm 0.37$} \\
VAE    & $97.78 \pm 0.08  $        & $86.90 \pm 0.17$       & $44.39 \pm 0.37$        
\end{tabular}
\end{table}

\begin{figure}[!t]
\centering
\includegraphics[width=0.825\linewidth]{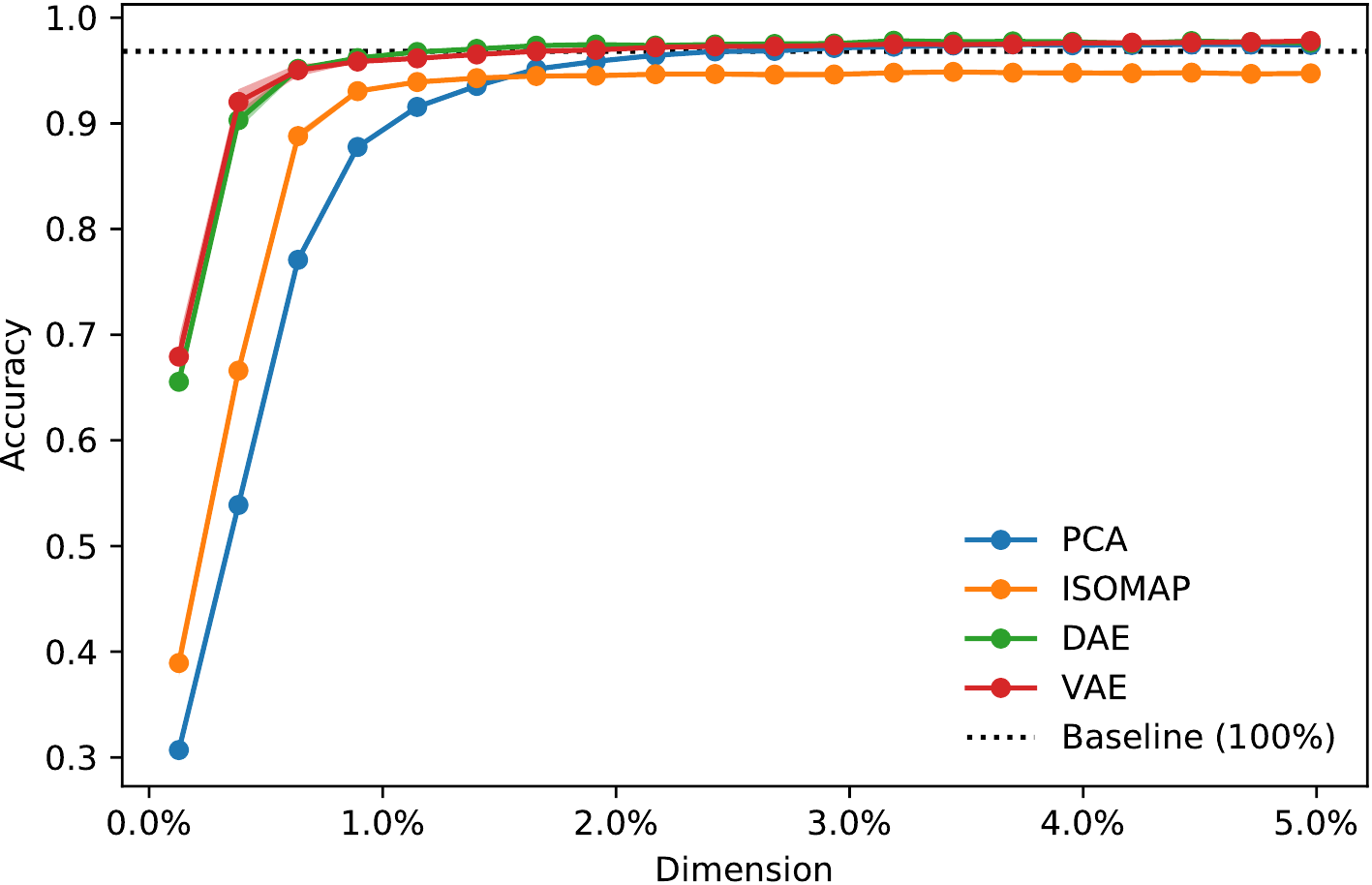}\\\vspace{5pt}
\includegraphics[width=0.825\linewidth]{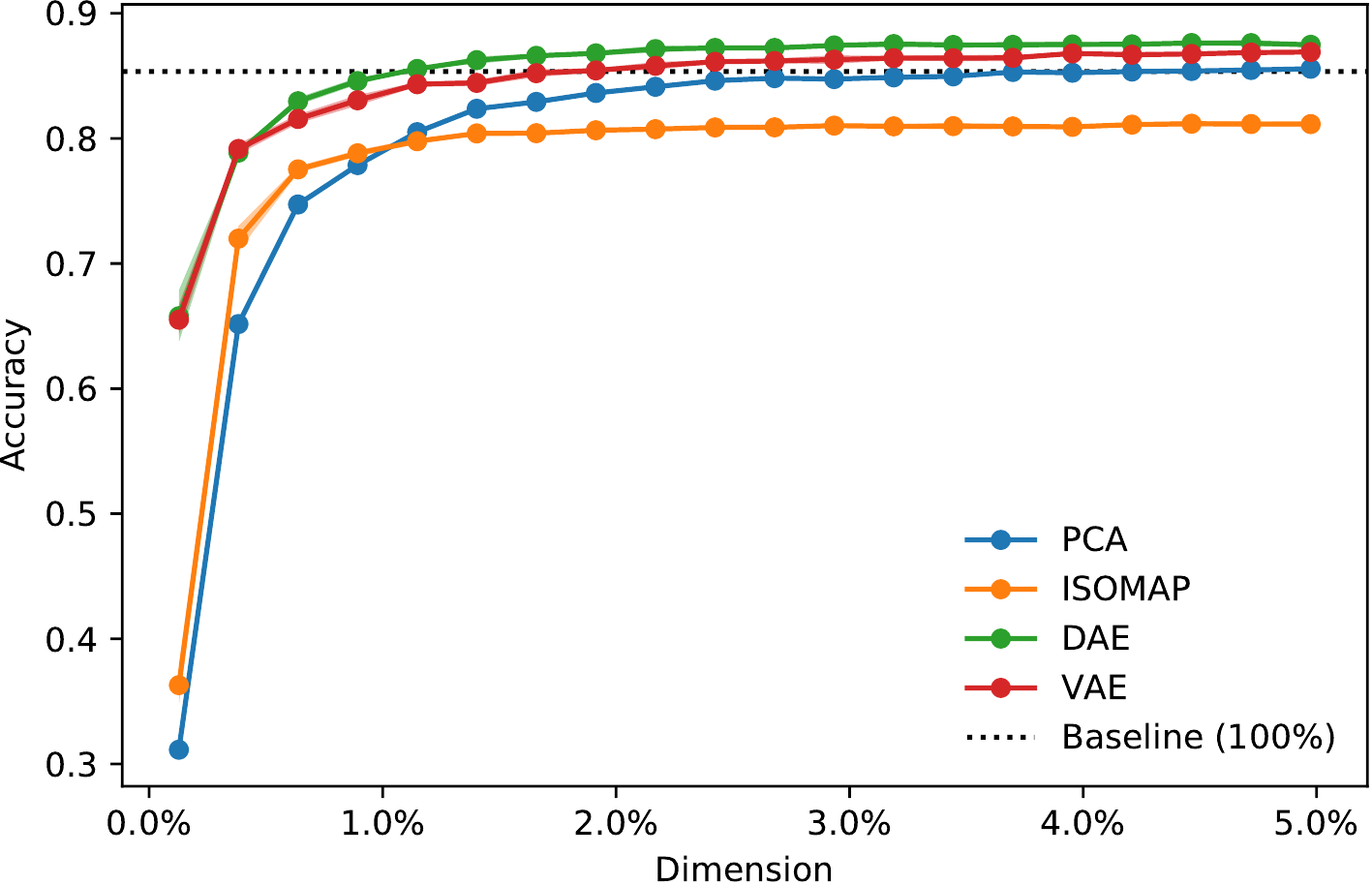}\\\vspace{5pt}
\includegraphics[width=0.825\linewidth]{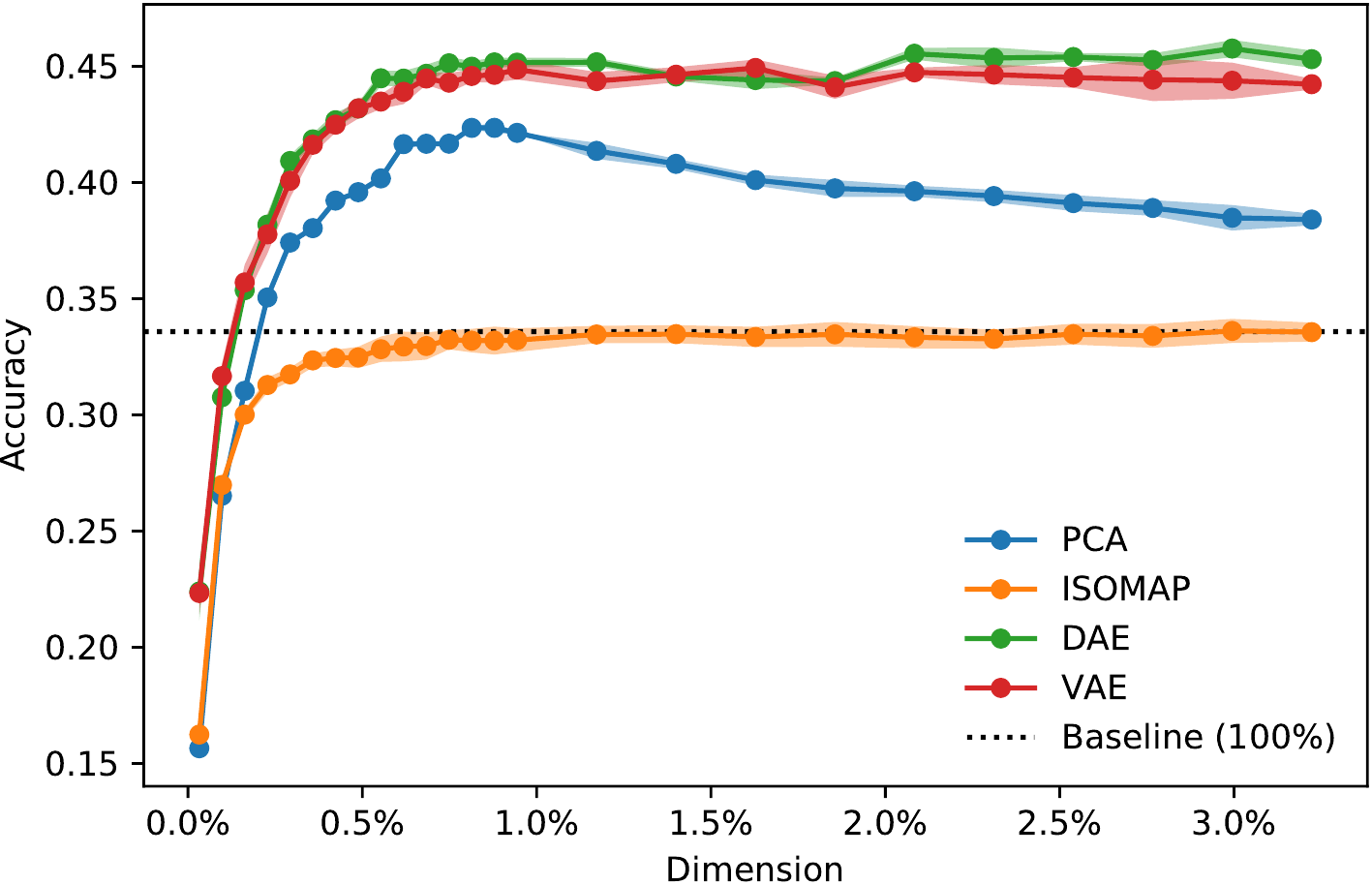}
\caption{$k$-NN mean accuracy with regards to the relative dimension of the projection. The shade indicates the standard deviation. The baseline is obtained by training $k$-NN on the original space ($100\%$ of the dimensions). From top to bottom row: MNIST, Fashion-MNIST, and CIFAR-10.}
\label{accuracy}
\end{figure}

Let us first consider the dimension's impact on $k$-NN mean accuracy (\figurename \ref{accuracy}). As expected, for small dimensions, PCA yielded the lowest accuracy as it is not able to grasp non-linearity, and consequently, to learn complex transformations. Isomap yielded slightly better accuracy than PCA as it is able to learn a projection that preserves the intrinsic structure of the data. Both neural networks consistently yielded the best accuracy as they are able to learn highly non-linear transformations. 

As we allow the projection's dimension to grow, the difference between each method reduces.
PCA and neural networks yielded comparable accuracy as it became simpler to learn a projection that preserves enough information for classification. The difference of accuracy between PCA and autoencoder best projection is only $0.34\pm0.08\%$, $2.06\pm0.09\%$ and $3.42\pm0.31\%$ on MNIST, Fashion-MNIST, and CIFAR-10 respectively. Asymptotically, Isomap consistently yielded the lowest accuracy. 

On CIFAR-10, there seems to be an optimal projection's size for PCA. While smaller dimensions do not hold sufficient information to correctly classify the data, it is likely that higher dimensions bring more noise than information, making the classification task harder. Note that $k$-NN is not well suited for CIFAR-10 -- hence its poor accuracy -- and should be replaced in practice. 

Note that accuracy's standard deviation is very low -- barely visible on the graphs -- which indicate that all methods repeatedly learned a similar projection in the context of classification.

\subsubsection{Computation Time}

\begin{figure}[!t]
\centering
\includegraphics[width=0.825\linewidth]{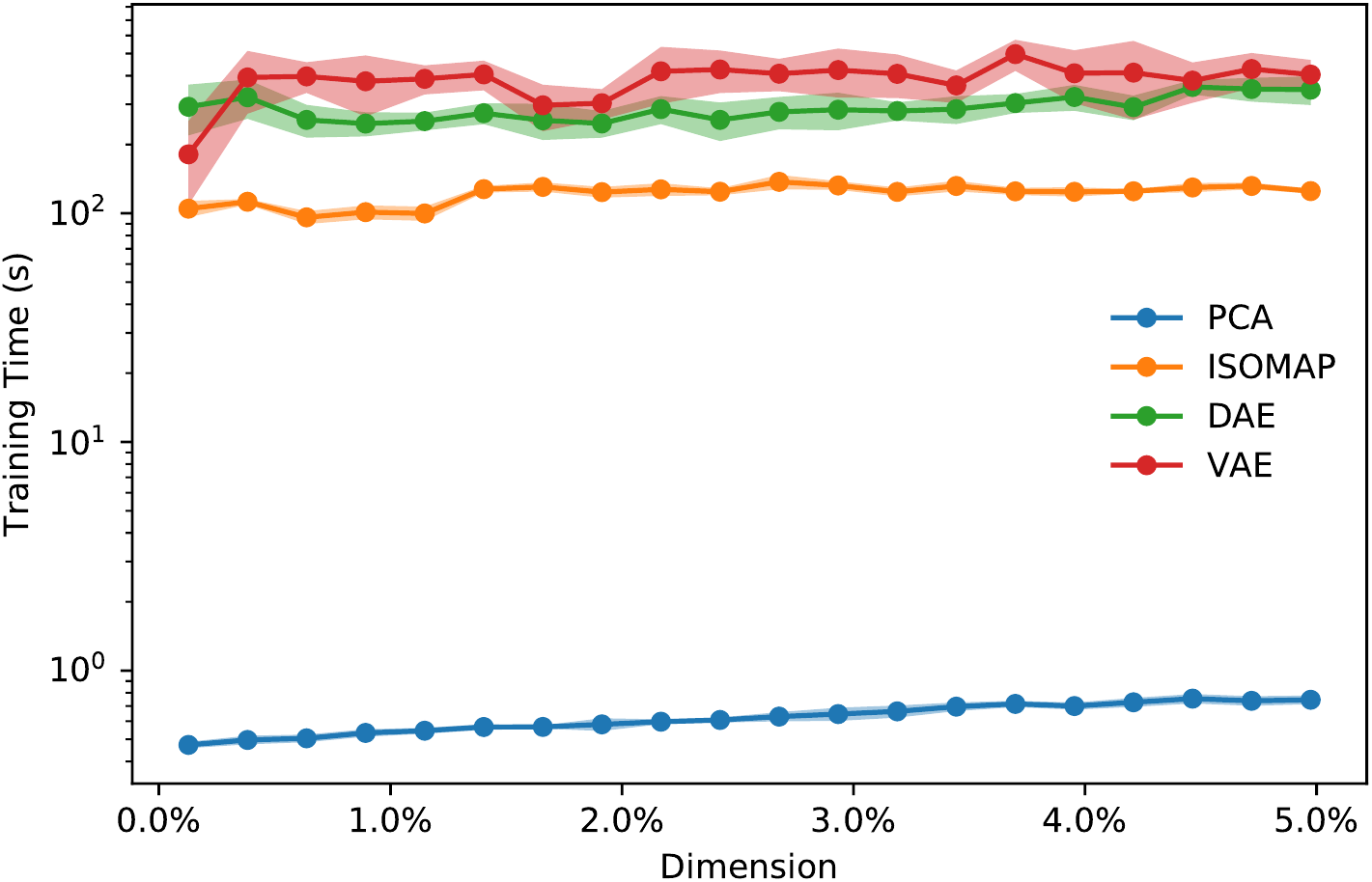}\\\vspace{5pt}
\includegraphics[width=0.825\linewidth]{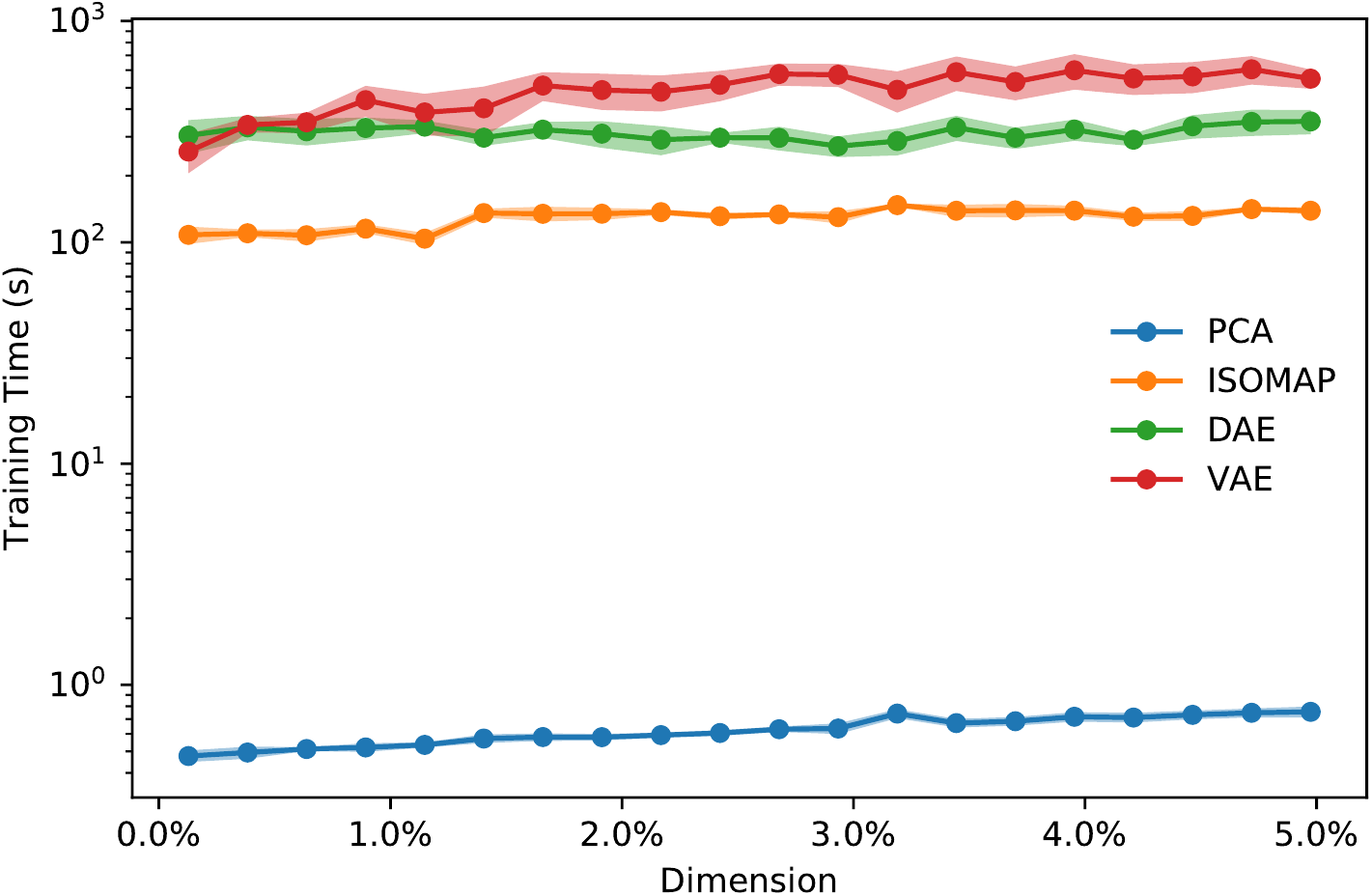}\\\vspace{5pt}
\includegraphics[width=0.825\linewidth]{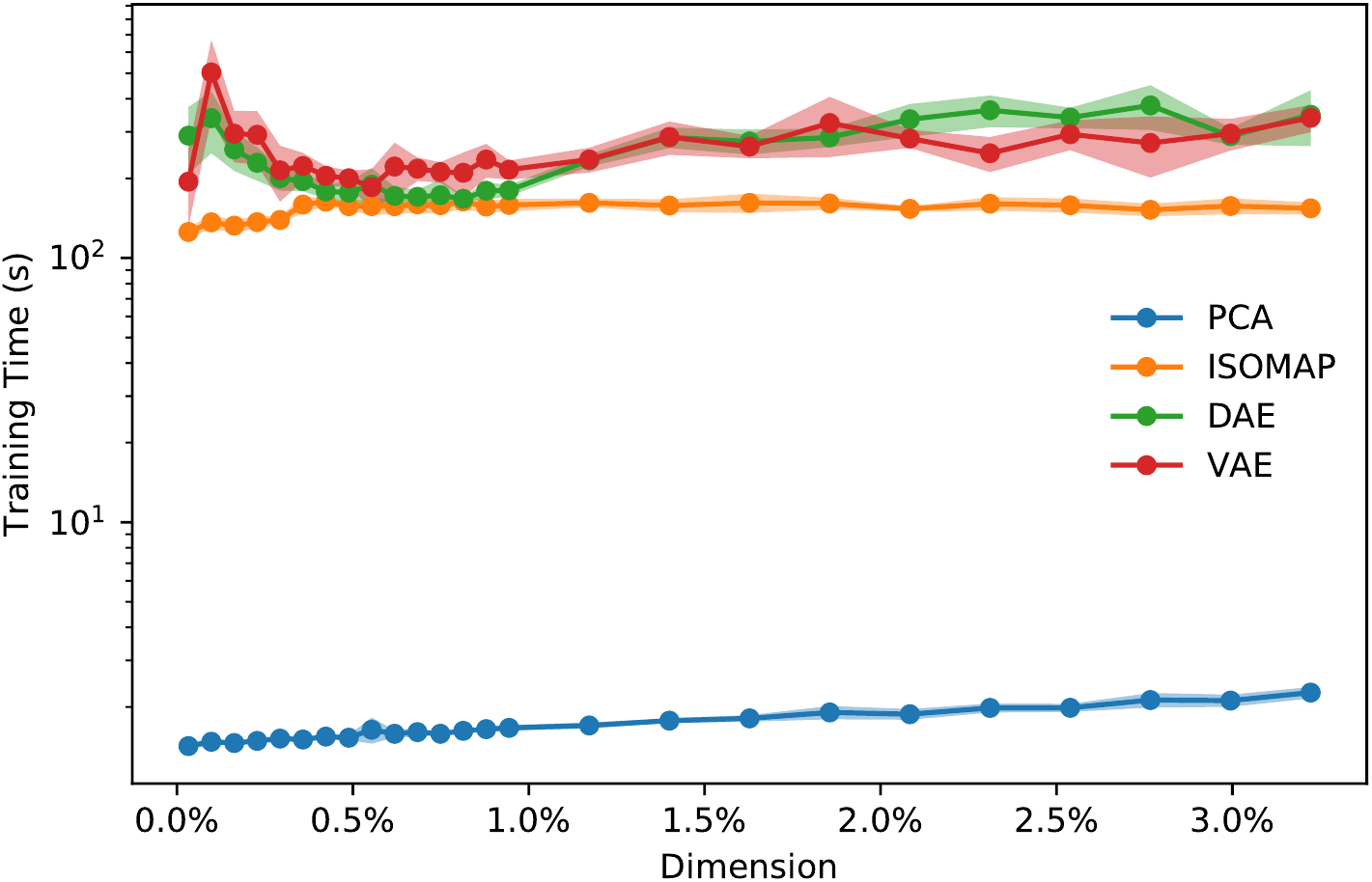}
\caption{Computation time in second with regards to the relative dimension of the projection. From top to bottom row: MNIST, Fashion-MNIST, and CIFAR-10.}
\label{time}
\end{figure}

Figure \ref{time} reports the average computation time of each projection method. Note that it does not include the $k$-NN classifier and that the scale is logarithmic.

The computation time depends very little on the dimension as SVD-based methods usually need to find every eigenvector\footnote{\textit{Scikit-learn} implementation of PCA use \textit{randomized SVD} introduced by Halko et al.~\cite{2009arXiv0909.4061H} which does not require to compute the complete eigendecomposition, hence the increase in computation time with the dimension.} and autoencoder's number of parameters vary very little. PCA, which is the simplest method to compute, was the fastest by two orders of magnitude. Isomap, whose number of training examples was capped at 10,000, took around 100 seconds to train. Compared to Isomap, DAE and VAE were about twice and four times as long to train, respectively.

Neural networks' training time depends to some extent on the number of epochs, which is controlled by early stopping. The average standard deviation of DAE and VAE computation time is $36.2$s ($12.8\%$) and $67.1$s ($18.2\%$), respectively. It can be challenging to estimate the computational time required to train a neural network, whereas PCA average standard deviation is only $0.04$s ($3.8\%$).

Even when there is an optimal dimension and that a random search over the size of PCA projection is needed, PCA is still faster. On CIFAR-10, it took $43\pm2$s to compute all PCA projections, whereas it took in average $249\pm32$s and $259\pm45$s to train a deep and a variational autoencoder, respectively.

One can reduce the computation time of neural networks by using more powerful graphic cards. Note that the Titan Xps used are high-end GPUs worth US\$ 1,200 each in 2018.

\section{Conclusion}
\label{conclusion}

On three commonly used image datasets, and in a context of $k$-NN classification, PCA allows for a comparable accuracy as autoencoders at a fraction of the computation time and resources. Supplementary material shows the same trend on both logistic regression and quadratic discriminant analysis. This observation holds true even if one needs to do a random search on PCA projection size. Furthermore, SVD-based methods do not require an expensive GPU. 


Our advice would be to use PCA while prototyping on large datasets to speed up computations, after which it can be replaced by a more flexible method such as an autoencoder to improve performance.

Following work should include a broader range of datasets.

\section*{Acknowledgment}
This research was financed by the Natural Sciences and Engineering Research Council of Canada (NSERC) under grant 2017-05617. This research was enabled in part by support provided by Calcul Quebec and Compute Canada.


\end{document}